\documentclass[11pt,a4paper]{article}
\usepackage[hyperref]{eacl2021}
\usepackage{times}
\usepackage{latexsym}

\usepackage{microtype}
\usepackage[utf8]{inputenc}
\usepackage[arabic,USenglish]{babel}
\usepackage{graphicx}
\aclfinalcopy 

\usepackage{float}
\restylefloat{table}

\title{A Multilingual African Embedding for FAQ Chatbots}

\author{Aymen Ben Elhaj Mabrouk \\ {\bf Moez Ben Haj Hmida} \\ {\bf Chayma Fourati} \\ {\bf Hatem Haddad} \\ {\bf Abir Messaoudi} \\ { iCompass, Tunisia} \\ \texttt{\{aymen,moez,chayma,hatem,abir\}@icompass.digital}}

\date{}

\begin{document}
\maketitle

\begin{abstract}

Searching for an available, reliable, official, and understandable information is not a trivial task due to scattered information across the internet, and the availability lack of governmental communication channels communicating with African dialects and languages. In this paper, we introduce an Artificial Intelligence Powered chatbot for crisis communication that would be omnichannel, multilingual and multi dialectal. We present our work on modified StarSpace embedding tailored for African dialects for the question-answering task along with the architecture of the proposed chatbot system and a description of the different layers. English, French, Arabic, Tunisian, Igbo,Yorùbá, and Hausa are used as languages and dialects. Quantitative and qualitative evaluation results are obtained for our real deployed Covid-19 chatbot.
Results show that users are satisfied and the conversation with the chatbot is meeting customer needs.
\end{abstract}

\section{Introduction}

Nowadays, African people have easier access to the internet and particularly to digital information through various platforms. However, African public and private institutions interact with citizens through old and non updated websites using the official language to communicate, mainly French or English. Nevertheless, most African citizens talk and express themselves in their own local dialects which makes the information not accessible and not easy to find. On the other hand, they are struggling with the geographic distance from the institutions.
In addition to that, African institutions phone numbers to be contacted are bombarded and unavailable especially during a crisis case such as the Covid-19 pandemic where everyone is asking for a piece of information about the pandemic or how to prevent it. Faced with the Covid-19 crisis, African institutions faced the need to improve the communication across digital channels and struggled to make the information as reliable as possible, in the official language and in local dialects.

In daily life, Tunisian people, for instance, communicate and type using the Tunisian dialect. It consists of ``Tunisian Arabic'' also described as ``Tounsi'' or ``Derja'' which uses arabic alphabets but is different from the Modern Standard Arabic (MSA) and ``TUNIZI'' which was introduced in~\cite{Chayma2020} as ``the Romanized alphabet used to transcribe informal Arabic for communication by the Tunisian Social Media community''. In fact, Tunisian dialect features Arabic vocabulary spiced with words and phrases from Tamazight, French, Turkish, Italian and other languages \cite{tounsi}. In \cite{younes}, a work on Tunisian Dialect on Social Media mentions that \textbf{81\%} of the Tunisian comments on Facebook used Romanized alphabet. A study was conducted in~\cite{abidi} on social media Tunisian comments presents statistics as follows: Out of 1,2M comments with 16M words and 1M unique words,  53\% of the comments use Romanized alphabet, 34\% use Arabic alphabet, and 13\% use script-switching. 87\% of the Romanized alphabet are Tunizi, while the rest are French and English.

A work\footnote{\url{https://translatorswithoutborders.org/language-data-nigeria/}} presents statistics about Nigerian languages and dialects and indicates that ``Nigeria is one of the most linguistically diverse countries in the world, with over 500 languages spoken''. Nigeria's official language is English, however is spoken less frequently in rural areas and amongst people with lower educational levels. Major languages spoken in Nigeria include Hausa, Yorùbá, Igbo, and Fulfulde. 
Statistics\footnote{\url{https://www.worlddata.info/languages/hausa.php}} indicates that a total of about 54.8 million people worldwide speak Hausa as their mother tongue.
Statistics\footnote{\url{https://nalrc.indiana.edu/doc/brochures/igbo.pdf}} are presented as follows: The "Igbo people number between 20 and 25 million, and they live all over Nigeria". An other work\footnote{\url{https://k-international.com/blog/most-spoken-african-languages}} indicates that the most spoken African languages by number of native speakers are Arabic, Berber (Amazigh), Hausa, Yorùbá, Oromo, Fulani, Amharic, and Igbo.
 
To answer properly a huge number of Covid-19 related questions, asked at the same time, we decided to develop a chatbot. However, two main problems were encountered. First, the non availability of covid-19 related lexical field in any of previously deployed chatbots. Also, no pretrained language models for dialects, particularly, African ones exist. Hence, the creation of our own specific domain embedding model is crucial.\\
In this paper, we introduce a chatbot for African institutions that is omnichannel, multilingual and multi-dialectal. This solution would solve the digital communication with citizens using official governmental language and local dialects, avoiding fake news, understanding and replying using the relevant language or dialect to the user. Services would be digitized and hence, illiterate, and vulnerable citizens such as rural women and people that need social inclusion and resilience would have easier access to reliable information.

\section{Related works}
Open-domain Question-Answering (QA) task has been intensively studied to evaluate the performances Language Understanding models. This task takes as input a textual question to look for correspondent answers within a large textual dataset. In \cite{abs-1906-05394}, two MSA QA datasets have been proposed.\\ In 2016, the first Arabic dialect chatbot,called BOTTA, was developed that uses the Egyptian Arabic dialect in the conversation \cite{abu}. It represents a female character that converses with users for entertainment. BOTTA was developed by using AIML and was launched by using Pandorabots platform.\\
In \cite{Al-Ghadhban2020}, authors developed a social chatbot that can support conversation with the students of the information technology (IT) department at King Saud University (KSU) using the Saudi Arabic dialect. They collected 248 inputs/outputs from the KSU IT students, then preprocessed and classified the collected data into several text files to build the dialogue dataset.

To the best of our knowledge, no such work was previously done for African dialects.

\section{Methodology}
In order to create our chatbot, three steps were conducted.
\begin{itemize}
    \item First, we collected data from official sources. For instance, Tunisian information was provided from Ministry of health and Nigerian information from Nigerian local Non-Governmental Organizations (NGOs). 

    \item Second, already collected data needs to get further augmented and spiced with chitchat like greetings and jokes.
    
    \item Finally, data collected gets divided into two main categories: Frequently Asked Questions (FAQ) and chitchat.

\end{itemize}

\subsection{Data Collection}
In order to collect Tunisian and Nigerian reliable and official data, covid-19 related questions and answers were provided by the Tunisian ministry of health, and Nigerian NGOs respectively.
However, such information are available only in official languages. For instance, Tunisian resources are written in French and Arabic, and Nigerian ones in English. Hence, the need to include local dialects spoken by citizens on a daily basis.

\subsection{Data Augmentation}

In order to further augment our data and include local dialects, each question written in French was translated into Tunisian dialect using the Tunizi way of writing. Questions collected in MSA are translated into Tunisian dialect expressed in Arabic letters.
Every question was augmented in Tunisian dialect by five Tunisian native speakers. 
Table \ref{Data Tunisian Samples} presents the translation of the question ``How to protect myself from the covid-19?'' into French, MSA, Tunizi, Tunisian Arabic, and English.
Answers for the Tunisian asked questions written in Tunizi, or French are answered in French, whereas questions written in Tunisian Arabic or MSA are answered in MSA.

Since Nigerian collected questions were expressed only in English, all questions and answers were translated into Igbo, Yorùbá, and Hausa respectively. This work was done by seven Nigerian native speakers. Hence, a question asked in one of the three dialects would be answered in the same chosen dialect. Table \ref{Data Nigerian Samples} presents the question ``What is the Current Status of COVID-19 Testing in Nigeria?'' expressed and answered in English and all available Nigerian dialects.

\begin{table*}[h!]
\centering
\begin{tabular}{p{0.2\textwidth}p{0.2\textwidth}p{0.3\textwidth}}
\hline \textbf{Language / Dialect} & \textbf{Questions} & \textbf{Answers} \\ \hline
French & comment se proteger du covid-19 ? & Il faut bien laver les mains, porter une bavette et respecter la distanciation physique \\
MSA & \foreignlanguage{arabic}{كيف أحمي نفسي من الكورونا؟} & \foreignlanguage{arabic}{ يجب أن تغسل يديك جيدًا وأن ترتدي كمامة وأن تحترم التباعد الجسدي  }  \\
Tunizi & kifech ne7mi rou7i mel corona ? & Il faut bien laver les mains, porter une bavette et respecter la distanciation physique\\
Tunisian Arabic & \foreignlanguage{arabic}{كيفاش نجم نحمي روحي مل كورونا؟}  & \foreignlanguage{arabic}{يجب أن تغسل يديك جيدًا وأن ترتدي كمامة وأن تحترم التباعد الجسدي  }   \\
English  & How to protect myself from covid-19 ? & You must wash your hands well, wear a bib and respect physical distancing  \\
\hline
\end{tabular}
\caption{\label{Data Tunisian Samples} Tunisian Data  Samples}
\end{table*}

\begin{table*}[h!]
\centering
\begin{tabular}{p{0.2\textwidth}p{0.2\textwidth}p{0.3\textwidth}}
\hline \textbf{Language / Dialect} & \textbf{Questions} & \textbf{Answers} \\ \hline
Yorùbá & kini ipo sise ayewo ajakale arun COVID-19 ni orile-ede Naijiria? & fun ekunrere alaye  leri ajakale arun covid-19 ni orile-ede Naijiria, jowo lo si https://covid19.ncdc.gov.ng/ \\
Hausa & Mene ne Matsayin gwaji na COVID-19 a yanzu a Najeriya? & Don karin bayani, a ziyarce https://covid19.ncdc.gov.ng/ \\
Igbo & Kedu onodu nyochaputa kovid – 19 no na ya na Niajiria  ugbu a & Maka inweta ngwata zuru oke banyere kovidi– 19 n’ala Niajiria  biko gaan’igwe nomba a :https://covid19.ncdc.gov.ng/ \\
English  & What is the Current Status of COVID-19 Testing in Nigeria? & For up-to-date information on the Covid-19 situation in Nigeria, Please visit https://covid19.ncdc.gov.ng/ \\
\hline
\end{tabular}
\caption{\label{Data Nigerian Samples} Nigerian Data  Samples}
\end{table*}

\subsection{Data Grouping}

After collecting and augmenting the needed data in all languages and dialects, we grouped it into two main categories presented as follows: 
\begin{itemize}
    \item \textbf{FAQs} containing all official and reliable covid-19 related questions and answers.
    \item \textbf{Chitchat} containing informal conversations like salutations, jokes, greetings, etc. Chitchat answers are presented in the local dialect of the asked question.
\end{itemize}

Table \ref{chitchat_tun} presents examples for Tunisian chitchat in MSA, French, Tunizi, and Tunisian Arabic.

Table \ref{chitchat_nig} presents examples for Nigerian chitchat in English, Yorùbá, Hausa, and Igbo.

\begin{table*}[h!]
\centering
\begin{tabular}{p{0.2\textwidth}p{0.2\textwidth}p{0.3\textwidth}}
\hline \textbf{Language / Dialect} & \textbf{Questions} & \textbf{Answers} \\ \hline
French & je suis ravi &  Moi de même, vous êtes les bienvenus.\\
MSA & \foreignlanguage{arabic}{  صباح الخير} & \foreignlanguage{arabic}{صباح النور}  \\
Tunizi & 3asslama & Mar7be bik \\
Tunisian Arabic & \foreignlanguage{arabic}{  نهارك زين  }  & \foreignlanguage{arabic}{ أهلا وسهلا بيك، كيفاش انجم نعاونك؟ }   \\
\hline
\end{tabular}
\caption{\label{chitchat_tun} Tunisian Chitchat Samples}
\end{table*}

\begin{table*}[h!]
\centering
\begin{tabular}{ccc}
\hline \textbf{Language/Dialect} & \textbf{Question} & \textbf{Answer} \\ \hline
English & how r u & fine thank you, how can I help you? \\
Yorùbá & Bawo ni? & Daada, kini mo le se fun e?\\
Hausa & ya kike & Ina lafiya, yaya zan iya taimaka muku?\\
Igbo & Keduka idi? & O dimma i  mela, kee ka m ga-eki nyere gi aka?  \\

\hline
\end{tabular}
\caption{\label{chitchat_nig} Nigerian Chitchat examples}
\end{table*}

The number of answers of each category in each language and each dialect are presented in Table \ref{stats}.

\begin{table}[h!]
\centering
\begin{tabular}{lcc}
\hline \textbf{Language/Dialect} & \textbf{\#FAQ} & \textbf{\#Chitchat} \\ \hline
MSA \& Darija & 65 & 30\\
French \& Tunizi& 74 & 50\\
English & 182 & 27\\
Yorùba & 183 & 26\\
Hausa & 140 & 21\\
Igbo & 183 & 27\\
\hline
\end{tabular}
\caption{\label{stats} Content Statistics}
\end{table}

\section{Proposed System}
\begin{figure}[h!]
\includegraphics[width=8cm]{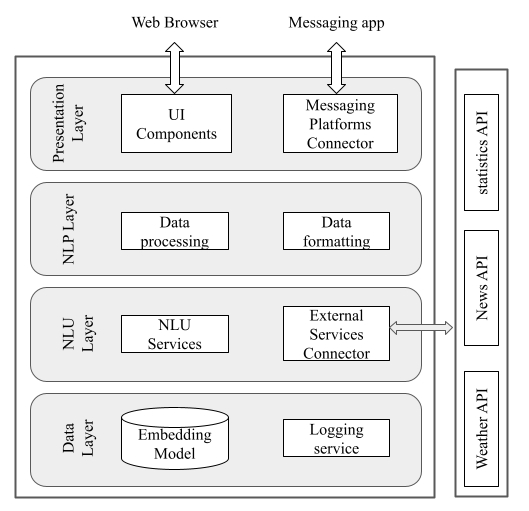}
\caption{Architecture of the proposed chatbot system.}
\label{workflow}
\centering
\end{figure}

 The proposed chatbot system is based on a multi-layered architecture as shown in Figure \ref{workflow}. 

The architecture  is  composed  of  four  layers: Presentation Layer, NLP Layer, NLU Layer, and Data Layer.
\begin{itemize}

    \item \textbf{The Presentation Layer} allows users to interact with the chatbot by posing questions and receiving responses. A simple web browser is enough to load the UI components and maintain a discussion with the chatbot. Alternatively, the interaction is also possible via messaging platforms. The messaging Platforms Connector provides built-in connectors to connect to common messaging platforms, like Facebook Messenger.

    \item \textbf{The NLP Layer} is composed of a data processing module and a data formatting module. The data processing module is responsible for data cleaning. The data formatting module receives HTML encoded questions from the presentation layer, then transforms them into the JSON (JavaScript Object Notation) format required by the NLU layer. Inversely, JSON responses received from the NLU layer are formatted in adequate HTML to be presented to users.

    \item \textbf{The NLU Layer} is considered to be the core of the chatbot system. This layer is composed of two modules: NLU services and External Services Connector. First, the NLU services module embeds the text question in a numerical vector. Second, the embedded question is mapped into an intent with a confidence score. If the score is higher than a predefined threshold, an answer related to the matched intent is determined. In specific cases, the answer requires an external service, like weather or news. In such a case, External Services Connector consumes an external API to retrieve the answer. When the confidence score is lower than the threshold, the question is considered ambiguous so it will be unanswered and is logged out by the lower layer. A default message is sent to the user stating that the chatbot has no specific answer for the question or asks the user to reformulate his/her question. The threshold is defined as :
   \[ \min_{q_{i} \in V{\cap}T}(\{C_{q_{i}}\})\]
 
 Where 
  \(V\) is the validation dataset,
  \(T\) is the set of all questions correctly predicted by the classifier, and 
  \(C_{q_{i}}\) is the confidence score of question \(q_{i}\).

    \item \textbf{The Data Layer} serves the NLU layer by storing the embedding model and providing logging service. This service furnishes tools to log unanswered questions in a file and log user conversations in a database for further analyses.

\end{itemize}

Since no previous work on Tunisian and Nigerian dialect Language models was previously done and specifically chatbots answering covid-19 questions, we trained a specific-domain language model dedicated for this task.

\subsection{Language Model}
We modeled our data as a document and a tag.
The documents represent the questions and the tags represent the intents. An intent defines the purpose of a user's question. 

Dialects, and particularly, African ones have no writing standards. People express themselves using words written in any format. For example the word ``good'' in Tunizi can be represented as ``behi'', ``bahi'', ``behii'' or ``behy'' etc. 
Therefore, we used bag of character n-grams representation for documents with n between two and four, and one hot encoding for tags.

\subsection{Classifier}
\begin{figure}[h!]
\includegraphics[width=8cm]{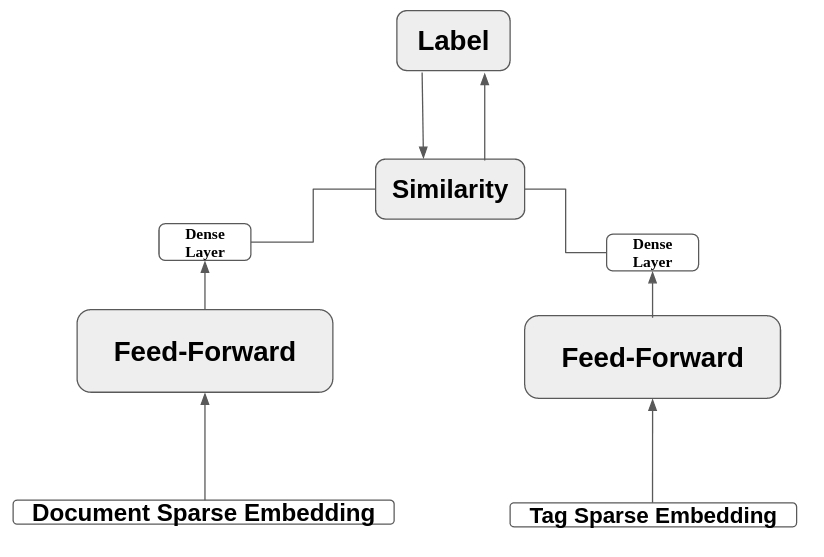}
\caption{Classifier}
\label{Classifier}
\centering
\end{figure}

We trained a multilingual and multidialectal classifier. We used StarSpace neural model\cite{StarSpace}. It embeds inputs and intent's labels into the same space and it is trained by maximizing similarity between them. We used two hidden layers and an embedding layer that returns a twenty dimensional vector. We used the default hyperparameters and the cosine similarity function. The classifier representation can be seen in Figure \ref{Classifier}.

\section{Results and discussion}
The chatbot deployment was done on two different channels: websites (official government website and NGO website) and Facebook Messenger (a Facebook page was created for the chatbot where users can interact with it using the Messenger). In the period from March 23, 2020 to June 18, 2020, 34800 users, excluding bot traffic, interacted with the chatbot as shown in Table \ref{Statistics}.

\begin{table}[H]
\centering
\begin{tabular}{lr}
\hline \textbf{Characteristic} & \textbf{Statistic} \\ \hline
Number of unique users & 34 800  \\
Number of received questions & 300 000  \\
Min  \#questions/conversation &  1\\
Max \#questions/conversation &  32\\
Average of questions/conversation & 7.4  \\
\hline
\end{tabular}
\caption{\label{Statistics} Chatbot Statistics from March 23, 2020 to June 18, 2020.}
\end{table}

\subsection{Quantitative evaluation}

For the Messenger channel, 8153 users interacted with our chatbot, with a minimum of 89 users/day and a maximum of 903 users/day.
 
The stickiness rate calculated as the ratio of Daily Active Users to Monthly Active Users achieved 16.85\%. 
Nevertheless, results show that, in a crisis context, users prefer to retrieve information from official websites rather than non-official ones like Messenger. 
In Table \ref{gender}, We present the distribution of Messenger's users over age and gender from April 1, 2020 to June 18, 2020.

\begin{table}[H]
\centering
\begin{tabular}{lrr}
\hline \textbf{Age} & \textbf{Female} & \textbf{Male} \\ \hline
13-17&10\%&7\%\\
18-24&13\%&16\%\\
25-34&12\%&18\%\\
35-44&12\%&18\%\\
45-54&2\%&3\%\\
55-64&1\%&2\%\\
65+&1\%&2\%\\

\hline
\end{tabular}
\caption{\label{gender} User distribution over age and gender.}
\end{table}

These statistics demonstrate the high interaction rate with our chatbot. We can conclude that people had real conversations where the average of messages during one conversation is at 7.4.

\subsection{Qualitative evaluation}
To measure the chatbot's quality, we used the Sensibleness and Specificity Average (SSA) metric. SSA is a human evaluation metric proposed by Google Brain Team \cite{ssa}. Sensibleness and Specificity Average measures human-likeness of the chatbot by answering the following question: Does the response have sense (Sensibleness) and was it specific (Specificity)?.

For example, if a user asks ``How to protect my-self from covid-19 ?'' and the chatbot responds with ``Be safe'', then the answer should be marked as ``not specific''. Such a reply could be used in various contexts. However, if the chatbot responds ``You must wash your hands well, wear a bib and respect physical distancing'', then the answer should be marked as ``specific'', since it relates closely to what is being discussed. However, the sensibleness metric is not sufficient when considered alone. Responses can be sensible but vague. Vague responses are not able to attract users to ask more questions. As an example, if the question is not Covid-19 related, the chatbot replies with a generic apology. Hence, the second dimension of the SSA metric will decide the specificity of the response.
\\
We asked Human judges to label answers as \textit{sensible}, when it has a meaning, and as \textit{specific} when it is close to the discussed topic. Answers labeled \textit{not sensible} are considered \textit{not specific}.
Sensibleness and specificity rates are calculated as the percentage of sensible-labeled and specific-labeled responses respectively.
We calculate the average of both metrics, which results in the SSA metric. SSA is a representative for human likeness, that also ignores chatbots with generic responses.\\

We randomly selected 100 discussions with 20 user interactions (questions). We submitted the discussions to 3 human judges and asked whether each response, given the context, is sensible and specific. We consider this evaluation as static since we have a fixed context: Covid-19. The obtained results are presented in Table \ref{ssa_table}.

\begin{table}[H]
\centering
\begin{tabular}{lrr}
\hline \textbf{Metric} & \textbf{Average} \\ \hline
Sensibleness & 68\%\\
Specificity &  60\% \\
\hline
\end{tabular}
\caption{\label{ssa_table} Average of SSA human evaluation of our chatbot.}
\end{table}

\section{Conclusion and Future work}
In this work, we present the first work for African chatbots that is covid-19 pandemic related. The particularity of this work is that this chatbot answers questions using English, French, Arabic, Tunisian, Igbo, Yorùbá, and Hausa without any predefined scenarios. Our solution digitalizes African institutions Services and hence, illiterate, and vulnerable citizens such as rural women and people that need social inclusion and resilience would have easier access to reliable, official, and understandable information 24/7.\\

The Artificial  Intelligence Powered chatbot for crisis communication is omnichannel,  multilingual and multidialectal. We modified StarSpace embedding to be tailored for African dialects along with its architecture. Quantitative and qualitative evaluations were performed. Results show that users are satisfied and that conversations with the chatbot are meeting customer needs. Indeed, statistics demonstrate the high interaction rate with the chatbot. SSA metric and its individual factors (specificity and sensibleness) showed reasonable performances.
The architecture used can be tailored to any tokenized African dialect. Hence, the ability to make other chatbots for other underrepresented African languages and dialects.\\
As a future work, we plan to build a voicebot that understands vocal requests and replies using voice messages. Such a problematic will be addressed using Speech To Text (STT) and Text To Speech (TTS) technologies for African dialects.

\bibliography{anthology,eacl2021}

\begin{thebibliography}{9}
\expandafter\ifx\csname natexlab\endcsname\relax\def\natexlab#1{#1}\fi

\bibitem[{Abidi(2019)}]{abidi}
Karima Abidi. 2019.
\newblock \href {https://tel.archives-ouvertes.fr/tel-02510812} {\emph{La
  construction automatique de ressources multilingues {\`{a}} partir des
  r{\'{e}}seaux sociaux : application aux donn{\'{e}}es dialectales du Maghreb.
  (Automatic building of multilingual resources from social networks :
  application to Maghrebi dialects)}}.
\newblock Ph.D. thesis, University of Lorraine, Nancy, France.

\bibitem[{Abu~Ali and Habash(2016)}]{abu}
Dana Abu~Ali and Nizar Habash. 2016.
\newblock \href {https://www.aclweb.org/anthology/C16-2044} {{B}otta: An
  {A}rabic dialect chatbot}.
\newblock In \emph{Proceedings of {COLING} 2016, the 26th International
  Conference on Computational Linguistics: System Demonstrations}, pages
  208--212, Osaka, Japan. The COLING 2016 Organizing Committee.

\bibitem[{Adiwardana et~al.(2020)Adiwardana, Luong, So, Hall, Fiedel,
  Thoppilan, Yang, Kulshreshtha, Nemade, Lu, and Le}]{ssa}
Daniel Adiwardana, Minh-Thang Luong, David~R. So, Jamie Hall, Noah Fiedel,
  Romal Thoppilan, Zi~Yang, Apoorv Kulshreshtha, Gaurav Nemade, Yifeng Lu, and
  Quoc~V. Le. 2020.
\newblock \href {http://arxiv.org/abs/2001.09977} {Towards a human-like
  open-domain chatbot}.

\bibitem[{Al-Ghadhban and Al-Twairesh(2020)}]{Al-Ghadhban2020}
Dana Al-Ghadhban and Nora Al-Twairesh. 2020.
\newblock \href {https://doi.org/10.14569/IJACSA.2020.0110357} {Nabiha: An
  arabic dialect chatbot}.
\newblock \emph{International Journal of Advanced Computer Science and
  Applications}, 11(3).

\bibitem[{Fourati et~al.(2020)Fourati, Messaoudi, and Haddad}]{Chayma2020}
Chayma Fourati, Abir Messaoudi, and Hatem Haddad. 2020.
\newblock \href {https://arxiv.org/submit/3091079} {Tunizi: a tunisian arabizi
  sentiment analysis dataset}.
\newblock In \emph{AfricaNLP Workshop, Putting Africa on the NLP Map. ICLR
  2020, Virtual Event}, volume arXiv:3091079.

\bibitem[{Mozannar et~al.(2019)Mozannar, Maamary, El~Hajal, and
  Hajj}]{abs-1906-05394}
Hussein Mozannar, Elie Maamary, Karl El~Hajal, and Hazem Hajj. 2019.
\newblock \href {https://doi.org/10.18653/v1/W19-4612} {Neural {A}rabic
  question answering}.
\newblock In \emph{Proceedings of the Fourth Arabic Natural Language Processing
  Workshop}, pages 108--118, Florence, Italy. Association for Computational
  Linguistics.

\bibitem[{Stevens(1983)}]{tounsi}
Paul~B. Stevens. 1983.
\newblock \href {https://doi.org/10.1080/01434632.1983.9994105} {Ambivalence,
  modernisation and language attitudes: French and arabic in tunisia}.
\newblock \emph{Journal of Multilingual and Multicultural Development},
  4(2-3):101--114.

\bibitem[{{Wu} et~al.(2017){Wu}, {Fisch}, {Chopra}, {Adams}, {Bordes}, and
  {Weston}}]{StarSpace}
L.~{Wu}, A.~{Fisch}, S.~{Chopra}, K.~{Adams}, A.~{Bordes}, and J.~{Weston}.
  2017.
\newblock Starspace: Embed all the things!
\newblock \emph{arXiv preprint arXiv:{1709.03856}}.

\bibitem[{Younes et~al.(2015)Younes, Achour, and Souissi}]{younes}
Jihen Younes, Hadhemi Achour, and Emna Souissi. 2015.
\newblock Constructing linguistic resources for the tunisian dialect using
  textual user-generated contents on the social web.
\newblock In \emph{Current Trends in Web Engineering}, pages 3--14, Cham.
  Springer International Publishing.

\end{thebibliography}
\bibliographystyle{acl_natbib}

\end{document}